\documentclass[10pt,twocolumn,letterpaper]{article}

\usepackage{iccv}
\usepackage{times}
\usepackage{epsfig}
\usepackage{graphicx}
\usepackage{amsmath}
\usepackage{amssymb}

\usepackage{multirow}
\usepackage{caption}
\usepackage{subcaption}
\usepackage{boldline}

\providecommand{\thickhline}{\hlineB{3}}
\providecommand{\xvar}{\mathbf{x}}
\providecommand{\zvar}{\mathbf{z}}
\providecommand{\wvar}{\mathbf{w}}
\providecommand{\cvar}{\mathbf{c}}

\usepackage[pagebackref=true,breaklinks=true,letterpaper=true,colorlinks,bookmarks=false]{hyperref}

\iccvfinalcopy 

\def\iccvPaperID{3535} 

\ificcvfinal\fi

\begin{document}

\title{Discond-VAE: Disentangling Continuous Factors from the Discrete}

\author{Jaewoong Choi\thanks{Equal contribution} \qquad Geonho Hwang\footnotemark[1] \qquad Myungjoo Kang\\
Seoul National University\\
{\tt\small \{chjw1475, hgh2134, mkang\}@snu.ac.kr}
}

\maketitle
\ificcvfinal\thispagestyle{empty}\fi

\begin{abstract}
In the real-world data, there are common variations shared by all classes (e.g\onedot category label) and exclusive variations of each class. We propose a variant of VAE capable of disentangling both of these variations. To represent these generative factors of data, we introduce two sets of continuous latent variables, \textit{private variable} and \textit{public variable}. Our proposed framework models the private variable as a Mixture of Gaussian and the public variable as a Gaussian, respectively. Each mode of the private variable is responsible for a class of the discrete variable.

Most of the previous attempts to integrate the discrete generative factors to disentanglement assume statistical independence between the continuous and discrete variables. Our proposed model, which we call Discond-VAE, DISentangles the class-dependent CONtinuous factors from the Discrete factors by introducing the private variables. The experiments show that Discond-VAE can discover the private and public factors from data. Moreover, even under the dataset with only public factors, Discond-VAE does not fail and adapts the private variables to represent the public factors.
\end{abstract}

\section{Introduction}
Learning disentangled representation of data without supervision has been considered as an important task for representation learning. \cite{BengioRepresentation} Although there are diverse quantitative measures for the disentangled representation \cite{BetaTCVAE, DCIMetric, BetaVAE, FactorVAE, DIPVAE, ModularityMetric}, most of the qualitative interpretation of disentanglement agrees on the statistical independence between each basic element of representation. In other words, each element of the disentangled representation corresponds to only one generative factor of data while being invariant to the others. Hence, the disentangled representation is naturally a concise and explainable feature of data. Various VAE-based models to obtain more disentangled representation in an unsupervised way have been proposed such as \cite{BetaTCVAE, JointVAE, DCIMetric, structured, BetaVAE, FactorVAE, bayesfactor, DIPVAE}.

In particular, JointVAE \cite{JointVAE} introduced discrete latent variables as well as continuous variables to represent the generative factors of data. For the real-world data, there are intrinsically discrete generative factors such as digit-type in the MNIST dataset. Therefore, it is natural to adopt a discrete variable to get a disentangled representation of those generative factors. However, JointVAE has a limitation of assuming the independence between the continuous and discrete variables. 

\begin{figure}[t]
\begin{center}
\includegraphics[width=0.8\linewidth]{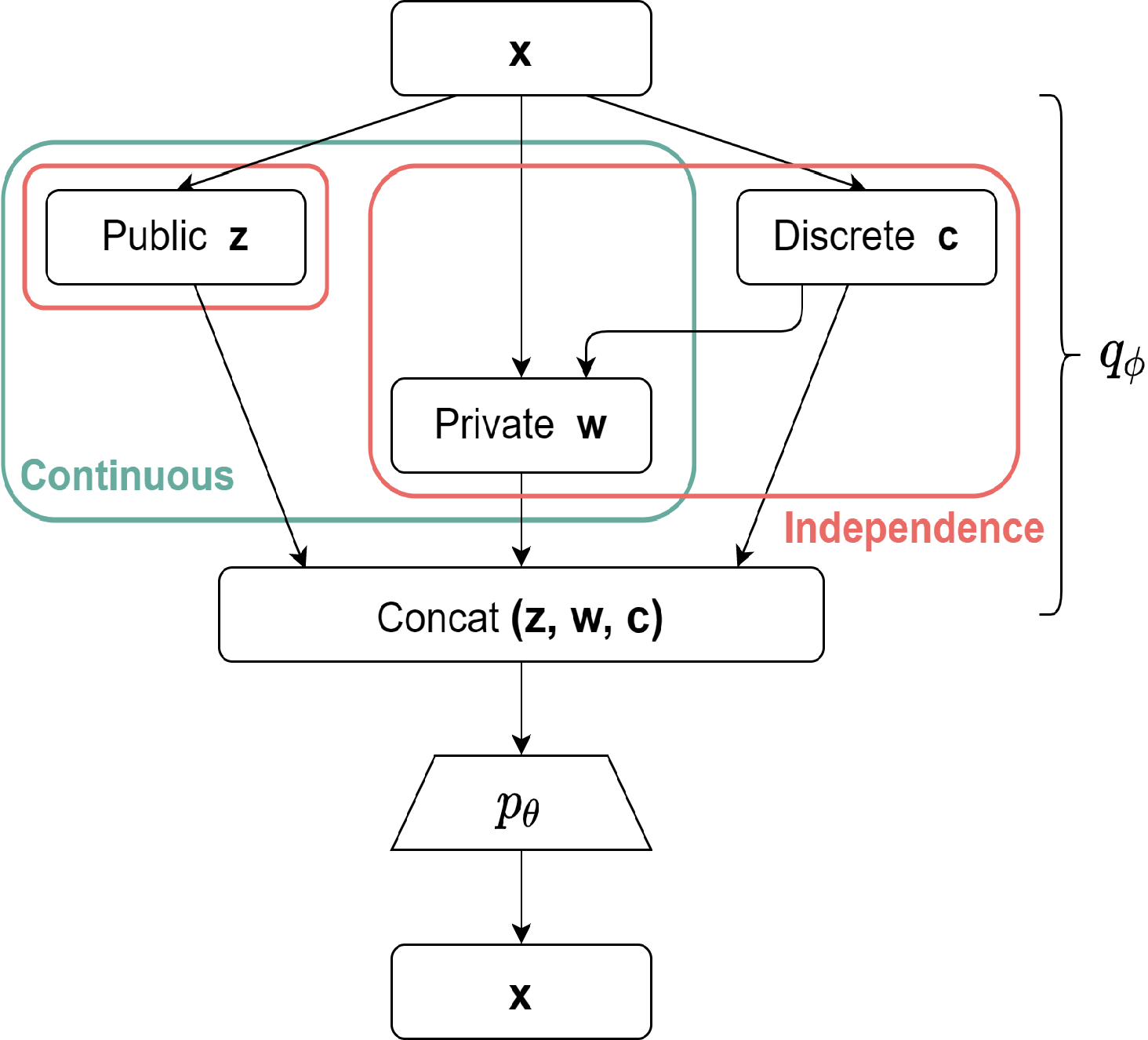}
\end{center}
\caption{Overview of Discond-VAE. Discond-VAE introduces two continuous latent variables (public and private variables) and one discrete variable to represent the data $\xvar$. The public continuous latent variable is assume to be independent to the private and discrete variables. 
}
\label{fig:model}
\end{figure}

The independence assumption of JointVAE is too restrictive to the real world. For example, consider the CelebA dataset \cite{CelebA}. The CelebA has 40 attribute labels, including Male(Gender) and Mustache. In this case, a continuous generative factor representing the Mustache volume is not independent with a discrete factor of the Gender. Hence, the class-independent continuous variable of JointVAE cannot properly represent the Mustache volume. From a similar perspective, \cite{disentangling} proposed a \textit{decomposition} to generalize the independence assumption of the disentanglement. \cite{disentangling} defines the decomposition of the latent variable as imposing the desired structure to the aggregate posterior $q_{\phi}(\zvar)$, such as sparsity or clustering. 

In this paper, we propose a new VAE model called Discond-VAE. Instead of imposing independence between the continuous and discrete factors, we propose learning the independent and dependent continuous factors jointly. Discond-VAE splits the continuous latent variable into two groups, \textit{private} and \textit{public} variable. We refer to each category of the discrete generative factor of data as \textit{class}. The \textit{private} variable represents variation within each class and the \textit{public} variable encodes the common generative factor of the entire classes. Therefore, Discond-VAE is able to represent the intra-class variation while keeping the capacity to represent the class-independent generative factor as in JointVAE.

Following the intuitive interpretation, we assume the public variable is independent of the discrete and private variables. The public and private variables are modeled by the Gaussian distribution and the Mixture of Gaussian, respectively. Each mode of the private variables corresponds to a class of the discrete variable. The experiments demonstrate that Discond-VAE can extract the private and public variables from data qualitatively and quantitatively.

\subsection{Contribution}
\begin{itemize}
    \item We propose a new VAE model called Discond-VAE. To the best of our knowledge, Discond-VAE is the first VAE model to represent the public and private continuous generative factors and the discrete generative factors at the same time.
    
    \item We propose a CondSprites dataset reassembled from the dSprites \cite{dSprites} to evaluate the disentanglement of private and public variables. The CondSprites dataset is designed to mimic the class-independent and class-dependent generative factors of the real-world.
    
    \item The existing disentanglement metrics assume the continuity of the latent variables and the independence of the generative factors. To integrate the discrete latent variable and class-dependent continuous variable into the disentanglement metrics, we propose a conditional disentanglement evaluation.

    \item We assess Discond-VAE on the CondSprites, dSprites, MNIST, and CelebA datasets. The experiments show that Discond-VAE can disentangle the public and private factors qualitatively and quantitatively.
    
\end{itemize}

\section{Background}
\subsection{VAE}
Variational Autoencoder (VAE, \cite{VAEKingma, VAERezende}) is a probabilistic model that learns a joint distribution $p(x, z)$ of the observed data $x$ and a continuous latent variable $z \in \mathbb{R}^{d}$. VAE models the joint distribution by
\begin{align}
    p(\zvar) &= \mathcal{N}(0, I_{d \times d}) \\
    p_{\theta}(\xvar \mid \zvar ) &= p(\xvar ; \mu_{\theta}(\zvar))
\end{align}
where $p(\xvar ;  \mu_{\theta}(\zvar))$ is a probabilistic distribution model with distribution parameters $\mu_{\theta} (\zvar)$. $p(\xvar ;  \mu_{\theta}(\zvar))$ is often referred to as the decoder. 

Given a dataset $\mathbf{X} = \{ \xvar_1, \xvar_2, \dots, \xvar_{N} \}$, VAE applies the variational inference by introducing an inference network $q_{\phi}(\zvar | \xvar)$, which is often referred to as the encoder. The encoder $q_{\phi}(\zvar | \xvar)$ approximates the true posterior $p_{\theta}(\zvar | \xvar)$ with a factorized Gaussian distribution with parameters encoded by the neural network. The encoder $q_{\phi}(\zvar | \xvar)$ and the decoder $p(\xvar ; \mu_{\theta}(\zvar))$ are simultaneously optimized by maximizing the Evidence Lower Bound (ELBO) $\mathcal{L}_{\textrm{VAE}}(\theta, \phi)$.
\begin{equation}
    \begin{split}
        \mathcal{L}_{\textrm{VAE}}(\theta, \phi) = \mathbb{E}_{q_{\phi}(\zvar \mid \xvar)} &\left[ \log p_{\theta}(\xvar \mid \zvar) \right] \\
        &-  D_{KL} (q_{\phi}(\zvar \mid \xvar) \mid\mid p(\zvar) ) 
    \end{split}
\end{equation}

The first term of the ELBO is the reconstruction loss which encourages the VAE to encode informative latent variables $\zvar$ to reconstruct the data $\xvar$. The second term regularizes the posterior distribution by promoting the encoded latent variables $q_{\phi}(\zvar | \xvar)$ to match the prior $p(\zvar)$. From this point of view, $\beta$-VAE \cite{BetaVAE} scales the regularization term of the ELBO by $\beta > 1$. The $\beta$-VAE induces more disentangled representations by matching the encoded latent variables with the factorized Gaussian prior $p(\zvar)$ by higher pressure \cite{VAECapacity, BetaVAE}.

\begin{equation} \label{eq:BetaVAE}
    \begin{split}
        \mathcal{L}_{\beta - \textrm{VAE}}(\theta, \phi) = \mathbb{E}_{q_{\phi}(\zvar \mid \xvar)} [ & \log p_{\theta}(\xvar \mid \zvar) ] \\
        - \beta \, &D_{KL} (q_{\phi}(\zvar \mid \xvar) \mid\mid p(\zvar) )    
    \end{split}
\end{equation}

\subsection{JointVAE}
VAE and $\beta$-VAE employs only continuous latent variables, especially factorized Gaussian, to model the latent variable $\zvar$. JointVAE \cite{JointVAE} generalizes the previous VAE and $\beta$-VAE by introducing discrete variables to disentangle the generative factors of the observed data.

Let $\zvar$ be a continuous latent variable and $\cvar$ be a discrete latent variable. By assuming conditional independence between the continuous and discrete latent variables, i.e. $q_{\phi}(\zvar, \cvar | \xvar) = q_{\phi}(\zvar | \xvar) \, q_{\phi}(\cvar | \xvar)$ and independent prior, i.e. $p(\zvar, \cvar) = p(\zvar) \, p(\cvar)$, JointVAE derived an optimization objective (Eq \ref{eq:JointVAE}) from the $\beta$-VAE objective (Eq \ref{eq:BetaVAE}). To prevent the posterior collapse phenomenon of the discrete latent variable, JointVAE applied the capacity control \cite{VAECapacity} to the objective.
\begin{multline} \label{eq:JointVAE}
    \mathcal{L}_{\textrm{Joint}}(\theta, \phi) = \mathbb{E}_{q(\zvar \mid \xvar, \cvar)} \left[ \log p(\xvar \mid \zvar) \right] \\
    - \beta_{\zvar} \, \mid D_{KL} (q(\zvar \mid \xvar) \, \mid\mid \, p(\zvar) ) - C_z \mid \\
    - \beta_{c} \, \mid D_{KL} (q(\cvar \mid \xvar) \, \mid\mid \, p(\cvar) ) - C_{c} \mid
\end{multline}

Since the sampling process from the categorical distribution is non-differentiable, the reparametrization trick can not be applied directly to the discrete encoder $q_{\phi}(\cvar | \xvar)$. To address this problem, JointVAE employed a differentiable relaxation of discrete variable sampling called Gumbel-Softmax distribution \cite{GumbelMax, GumbelSoftmax, ConcreteDist}. The Gumbel-Softmax provides a continuous approximation of the categorical distribution.

\section{Discond-VAE}
In this section, we describe the motivation and probabilistic formulation of Discond-VAE. Then, we describe how the probabilistic formulation is instantiated by a neural network.

\subsection{Motivation} \label{sec:motivation}
Although JointVAE \cite{JointVAE} extends the capability of VAE to encode discrete factors, JointVAE has a limitation of assuming the independence between continuous and discrete variables. This assumption usually does not hold for the general dataset. (\eg CelebA \cite{CelebA} or ImageNet \cite{ImageNet}). Therefore, we generalize the assumption.

Consider a generative modeling problem with the observed data $\xvar$, the discrete generative factor $\cvar$, and a set of continuous factors. Some of the continuous factors may be independent of the discrete factor $\cvar$, but the others may not. We refer to the former independent continuous factor as a \textit{public generative factor} and the latent variable representing it as a \textit{public variable $\zvar$}. Likewise, we call the latter dependent continuous factor as a \textit{private factor} and the latent variable representing it as a \textit{private variable $\wvar$}.

For example, consider a synthetic dataset of 2D Square and Ellipse images. For each shape, the images vary in scale and orientation. Also, the Square images vary in $x$-position and the Ellipse images vary in $y$-position. In this dataset, a latent variable that encodes scale and orientation should be independent of the discrete variable which encodes shape. However, the latent variables representing $x, y$-position should be dependent on the discrete variable. In short, this dataset has the public factor of scale and orientation, and the private factor of $x, y$-position. We refer to this dataset as CondSprites and use it to evaluate Discond-VAE in Sec \ref{Sec:CondSprites}. 

\subsection{Model}
\subsubsection{Probabilistic Model}
We propose a modification to JointVAE \cite{JointVAE} whose latent variable is composed of the discrete, public, and private variables. Since the public variable represents generative factors shared by all classes and the private variable represents variation within each class, it is natural to assume that the prior $p(\zvar, \wvar, \cvar)$ factorizes to $p(\zvar)$ and $p(\wvar, \cvar)$.
\begin{equation}
    p(\zvar, \wvar, \cvar)  = p(\zvar) \cdot p(\wvar, \cvar) = p(\zvar) \cdot p(\cvar) \cdot p(\wvar \mid \cvar)
\end{equation}
Likewise, the variational distribution $q_{\phi}(\zvar, \wvar, \cvar | \xvar)$ is modeled as the following.
\begin{equation}
\begin{split}
    q_{\phi} \left(\zvar, \wvar, \cvar \mid \xvar \right) = \,\, &q_{\phi}\left(\zvar \mid \xvar \right) \cdot q_{\phi}\left(\wvar, \cvar \mid \xvar \right) \\
     = q_{\phi}&\left(\zvar \mid \xvar \right) \cdot \,\, q_{\phi}( \cvar \mid \xvar ) \cdot q_{\phi}\left(\wvar \mid \xvar, \cvar \right)    
\end{split}
\end{equation}

For our Discond-VAE model, the $\beta$-VAE objective (Eq \ref{eq:BetaVAE}) becomes
\begin{multline} \label{eq:Discond-VAE}
    \mathcal{L}_{\textrm{Cond}}(\theta, \phi) = \mathbb{E}_{q_{\phi}(\zvar, \wvar, \cvar \mid \xvar)}[ \log p_\theta (\xvar \mid \zvar, \wvar, \cvar)] \\
         - \beta \, D_{KL}(q_{\phi}(\zvar, \wvar, \cvar \mid \xvar) \mid\mid p(\zvar, \wvar, \cvar))
\end{multline} 
The former log-likelihood term stands for the reconstruction error as in the previous VAE models. The latter KL divergence regularizer can be decomposed by the independence assumption.
\begin{equation}
      \begin{split}
     & D_{KL}(q_{\phi}(\zvar, \wvar, \cvar\mid\xvar) \mid\mid p(\zvar, \wvar, \cvar)
      \\
      &= D_{KL}(q_{\phi}(\zvar\mid\xvar) \mid\mid p(\zvar) ) + D_{KL}(q_{\phi}(\wvar, \cvar\mid\xvar) \mid\mid p(\wvar, \cvar))
    \end{split}
\end{equation}
Then, we can address the latter KL divergence as the following. (See appendix for proof)
\begin{equation}
    \begin{split}    
    D_{KL} (q_{\phi} (\mathbf{w}, \mathbf{c} &\mid \xvar) \mid\mid p(\mathbf{w}, \mathbf{c})) \\
    =\,  \mathbb{E}_{q_{\phi}(\mathbf{c}\mid \mathbf{x})} 
    & \left[ D_{KL} (q_{\phi}(\wvar \mid \xvar , \cvar) \mid\mid p(\wvar \mid \cvar)) \right] \\
    &+ D_{KL}(q_{\phi}(\mathbf{c} \mid \xvar) \mid\mid p(\mathbf{c}))
    \end{split}
\end{equation}
In brief, the learning objective of Discond-VAE (Eq \ref{eq:Discond-VAE}) is expressed as
\begin{equation}
    \begin{split}
         \max_{\theta, \phi} \,& \mathcal{L}_{\textrm{Cond}}(\theta, \phi) =\mathbb{E}_{q_{\phi}(\mathbf{z}, \mathbf{w}, \mathbf{c} \mid \mathbf{x})}[\log p_\theta (\mathbf{x} \mid\mathbf{z}, \mathbf{w}, \mathbf{c} )] \\
        &- \beta_{\zvar} \cdot D_{KL}(q_{\phi}(\mathbf{z} \mid \mathbf{x}) \mid\mid p(\mathbf{z}) ) \\
        &- \beta_{\wvar} \cdot \mathbb{E}_{q_{\phi}(\mathbf{c}\mid \mathbf{x})}
        \left [ D_{KL} (q_{\phi}(\wvar \mid \xvar , \cvar) \mid\mid p(\wvar \mid \cvar)) \right] \\
        & -\beta_{\cvar} \cdot D_{KL}(q_{\phi}(\mathbf{c} \mid \xvar) \mid\mid p(\mathbf{c}))
    \end{split}
\end{equation}

Discond-VAE models the $q_{\phi}\left(\zvar | \xvar \right), q_{\phi}( \cvar | \xvar )$ by the factorized Gaussian and Gumbel-Softmax as in the JointVAE \cite{JointVAE}. Moreover, Discond-VAE introduces the Gaussian Mixture encoder to model the joint distribution of the private and discrete variables. Each mode of the Mixture represents the generative factors within a class.
\begin{align} \label{eq:MixturePrivate}
    p(\wvar \mid \cvar) &= \prod_{i} p(\wvar \mid \cvar = e_{i})^{c_i}  \\
    &= \prod_{i} \mathcal{N}\left(\mu_i, I\right)^{c_i} \\
    q_{\phi}(\wvar \mid \xvar, \cvar) &= \prod_{i} q_{\phi}(\wvar \mid \xvar, \cvar=e_{i})^{c_i} \\
                        &= \prod_{i} \mathcal{N}\left(\mu_{i}(\xvar, \cvar), \Sigma_{i}(\xvar,\cvar)\right)^{c_i}
\end{align}
where $\cvar = ( c_{1}, c_{2}, \cdots, c_{d} ) \in \{0, 1\}^{d}$ denote a one-hot sample from the $d$-dimensional categorical distribution and $e_{i}$ denote a one-hot vector with $i$th component is one. Then, the KL divergence term of private variable $\wvar$ becomes

\begin{equation} \label{eq:KL_private}
    \begin{split}
        &\mathbb{E}_{q_{\phi}(\mathbf{c}\mid \mathbf{x})} 
        \left [ D_{KL} (q_{\phi}(\wvar \mid \xvar , \cvar) \mid\mid p(\wvar \mid \cvar)) \right] \\
        = &\sum_{i=1}^{d} \alpha_{i} \cdot D_{KL} (q_{\phi}(\wvar \mid \xvar , \cvar = e_{i}) \mid\mid p(\wvar \mid \cvar = e_{i}))
    \end{split}
\end{equation}
where $q_{\phi}(\mathbf{c}|\mathbf{x}) = (\alpha_1, \alpha_2, \cdots, \alpha_d)$ denotes the variational distribution of the discrete variable. Eq \ref{eq:KL_private} shows that the Discond-VAE encourages the disentanglement of private variables by regularizing the KL divergence to the prior by each mode. 

\subsubsection{Implementation}
We propose two methods to implement the probabilistic model of Discond-VAE. These two methods differ in how they encode and reparametrize the private variable $\wvar$.

\begin{figure}[!]
\begin{center}
\includegraphics[width=1\linewidth]{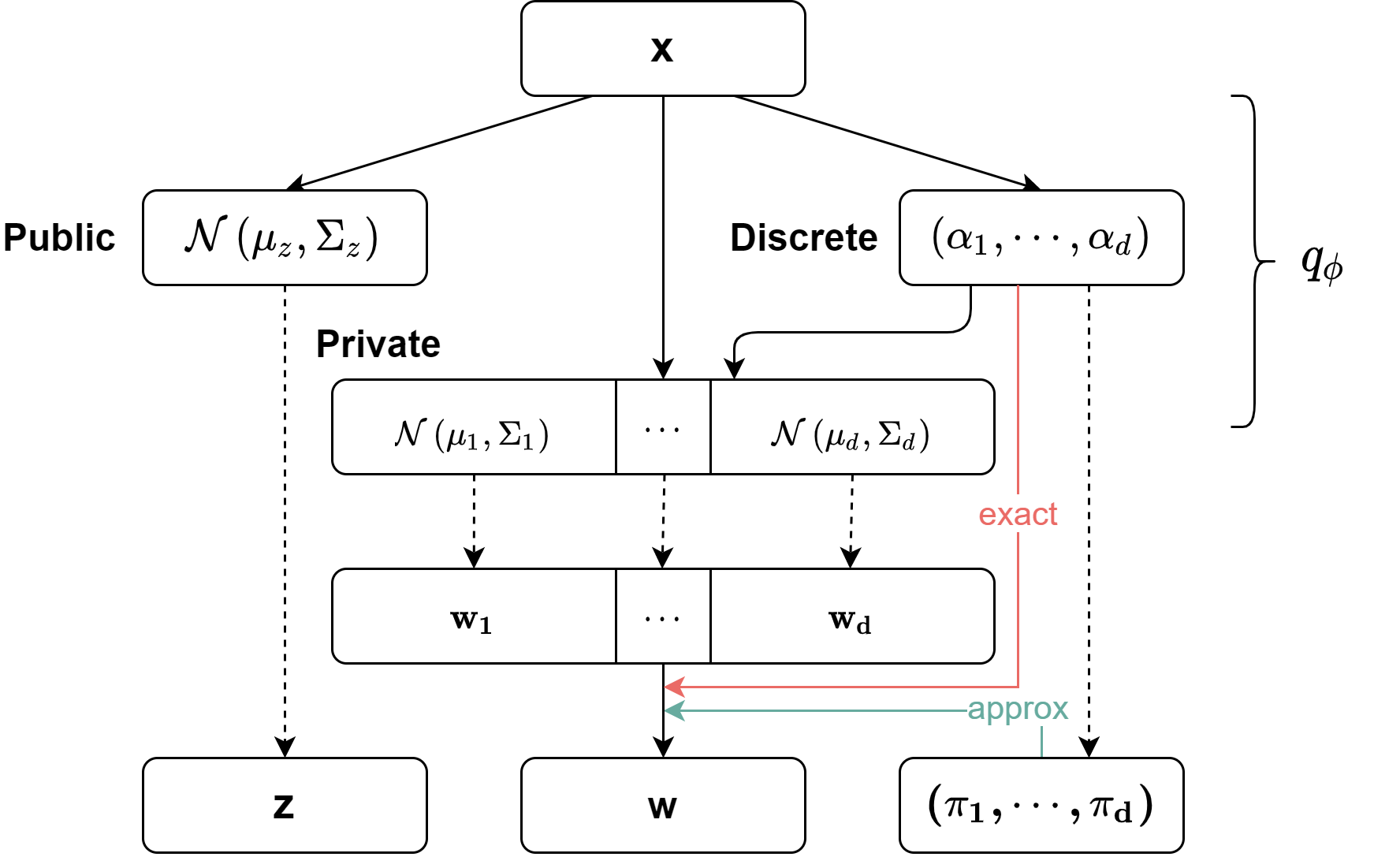}
\end{center}
\caption{Encoder and Sampling of Discond-VAE. The dashed lines denote the sampling from each variational distribution. The Discond-VAE-exact takes the private variable sample from the Mixture mode of the most likely class. The Discond-VAE-approx takes a linear combination of the Gaussian samples from each mode as the private variable sample.
}
\label{fig:reparametrization}
\end{figure}

First, we can model the posterior distribution $q_{\phi}(\wvar | \xvar, \cvar)$ while keeping the discreteness of the categorical variables.
For each class $\cvar = e_{i}$, the private variable encoder $q_{\phi}(\wvar | \xvar, \cvar)$ takes a concatenation of features extracted from the data $\xvar$ and \textit{one-hot encoding of the class} $e_{i}$ to infer the corresponding mode of the Gaussian Mixture.
\begin{equation}
    q_{\phi}(\wvar \mid \xvar, \cvar=e_{i}) = 
    \mathcal{N}\left(\mu_{i}(\xvar, e_{i}), \Sigma_{i}(\xvar, e_{i})\right)
\end{equation}
Note that the private variable encoder infers $d$-times where $d$ denotes the number of classes. For the reparametrization trick, this method takes a sample from the mode of the most likely class in $q_{\phi}(\mathbf{c}|\mathbf{x})$.
\begin{equation} \label{eq:ReparamPrivate_exact}
    \wvar = \wvar_{j}
\end{equation}
where  $j = \arg \max_{i} q_{\phi}(\mathbf{c} = e_{i} | \mathbf{x})$ and $\wvar_{j} \sim q_{\phi}(\wvar | \xvar, \cvar = e_{j})$. (Fig \ref{fig:reparametrization})  We refer to this model by \textit{Discond-VAE-exact}. Note that for the Discond-VAE-exact model with the perfect classification, the continuous variable encoder is a combination of the vanilla encoder applied to the entire dataset and the class-specific vanilla encoder applied only to the corresponding class.

Instead, the private variable encoder $q_{\phi}(\wvar | \xvar, \cvar)$ can take the data $\xvar$ and \textit{ the discrete variable} $q_{\phi}(\cvar | \xvar) = \boldsymbol{\alpha}$ to encode the private variable. We refer to this model by \textit{Discond-VAE-approx}.
\begin{equation}
    q_{\phi}(\wvar \mid \xvar, \cvar=e_{i}) = 
    \mathcal{N}\left(\mu_{i}(\xvar, \boldsymbol{\alpha}), \Sigma_{i}(\xvar, \boldsymbol{\alpha})\right)
\end{equation}
The Discond-VAE-approx model adopts a continuous approximation to the sampling from Gaussian Mixture $q_{\phi}(\wvar, \cvar | \xvar)$ by taking a linear combination of the samples from each mode $q_{\phi}(\wvar | \xvar, \cvar = e_{i})$. (Fig \ref{fig:reparametrization})
\begin{equation} \label{eq:ReparamPrivate_approx}
    \wvar = \sum_{i}^{d} \pi_{i} \cdot \wvar_{i}
\end{equation}
where $\wvar_{i} \sim q_{\phi}(\wvar | \xvar, \cvar = e_{i})$ and $\boldsymbol{\pi} = (\pi_{1}, \cdots, \pi_{d})$ denotes a sample from the Gumbel-Softmax distribution. As in the Discond-VAE-exact, the Discond-VAE-approx with the perfect classification of \textit{confidence $100\%$} has a continuous variable encoder equivalent to a combination of the public vanilla encoder and the class-specific vanilla encoder.

Two Discond-VAE implementations are optimized with the capacity objective \cite{VAECapacity} as in JointVAE to prevent a discrete variable from posterior collapsing. Hence, the learning objective of Discond-VAE becomes
\begin{equation}
    \begin{split}
        & \max_{\theta, \phi} \mathcal{L}_{\textrm{Cond}}(\theta, \phi) =\mathbb{E}_{q_{\phi}(\mathbf{z}, \mathbf{w}, \mathbf{c} \mid \mathbf{x})}[\log p_\theta (\mathbf{x} \mid\mathbf{z}, \mathbf{w}, \mathbf{c} )] \\
        &- \beta_{\zvar} \left| D_{KL}(q_{\phi}(\mathbf{z} \mid \mathbf{x}) \mid\mid p(\mathbf{z}) )  - C_{\zvar} \right|\\
        &- \beta_{\wvar} \left| \mathbb{E}_{q_{\phi}(\mathbf{c}\mid \mathbf{x})} 
        \left [ D_{KL} (q_{\phi}(\wvar \mid \xvar , \cvar) \mid\mid p(\wvar \mid \cvar)) \right]  - C_{\wvar} \right|\\
        & -\beta_{\cvar} \left| D_{KL}(q_{\phi}(\mathbf{c} \mid \xvar) \mid\mid p(\mathbf{c})) - C_{\cvar} \right|
    \end{split}
\end{equation}

The mean vector of each mode in Mixture prior $\mu_{i}$ are hyperparameters. The Discond-VAE-approx model sets the $\mu_{i}$ by the random samples from the standard Gaussian $\mathcal{N}(0, 1)$. Interestingly, the Discond-VAE-exact models show similar performance for the $\mu_{i} = 0$ with a smaller variance. Therefore, we set the $\mu_{i} = 0$ for the Discond-VAE-exact models in Sec \ref{Sec:experimetns}. We tried the EM algorithm or Warm-up approach to optimize the $\mu_{i}$. However, these approaches showed inferior performance and often unstable training dynamics. (The details of these approaches are provided in the appendix.)

\section{Experiments} \label{Sec:experimetns}
We evaluate the Discond-VAE model on the CondSprites, dSprites \cite{dSprites}, MNIST \cite{MNIST}, and CelebA \cite{CelebA} datasets. In the CondSprites, we compare the disentangling ability for the dataset with public and private generative factors. By contrast, the experiments on the dSprites evaluate the disentangling ability for the dataset with public generative factors only. Lastly, we evaluate the Discond-VAE model on the real-world dataset, MNIST, and CelebA, quantitatively and qualitatively.

For each dataset, we assess the Discond-VAE model in an unsupervised manner. Since the Discond-VAE divides continuous variables of the JointVAE into the private and public variables, we compare the Discond-VAE models with the same continuous dimension and the models with the same public dimension for a fair evaluation. For each quantitative score, we evaluate the model ten times randomly and report the means, standard deviations, and best scores. (See appendix for the full architecture and training hyperparameters.)


\begin{table}[!]
\centering
\begin{tabular}{l c c}
\thickhline
\multirow{2}{*}{Generative factors}      & \multicolumn{2}{c}{Shape} \\ \cline{2-3}
                        & Square & Ellipse\\ \hlineB{2}
Scale                   & \checkmark     & \checkmark     \\ 
Orientation             & \checkmark     & \checkmark     \\  
Position X              & \checkmark     &      \\  
Position Y              &      & \checkmark    \\ \thickhline
\end{tabular}
\caption{
Generative factors of the CondSprites.
}
\label{table:CondSprites_dataset}
\end{table}

\subsection{Dataset} \label{Sec:Dataset}
Since the MNIST and CelebA are standard benchmark datasets, we skip a detailed description of the datasets. For the CelebA, we center-cropped and resized each image to $64 \times 64$ resolution as in JointVAE \cite{JointVAE} following the custom of \cite{celebAresize}.

The dSprites \cite{dSprites} is a synthetic dataset to evaluate the disentanglement property of a model. Each sample is a 2D shape image generated from five generative factors. The dSprites dataset has one discrete generative factor of shape (square, ellipse, heart), and four continuous generative factors of scale, orientation, and position in $x, y$ axis. Since the dSprites assumes the independent generative factors, each combination of generative factors corresponds to an image. Thus, the dSprites has 737,280 images in total.

To evaluate disentangling ability further, we constructed a CondSprites dataset from the dSprites \cite{dSprites}. The CondSprites is designed to mimic the coexistence of class-independent variation and intra-class variation generative factors of the real-world. (See Sec \ref{sec:motivation} and Table \ref{table:CondSprites_dataset} for details) The CondSprites has 15,360 two-dimensional images consisting of 7,680 for Square and Ellipse, respectively.



\subsection{Quantitative Disentanglement Evaluation} \label{Sec:EvaluationMetric}
As a quantitative evaluation, we assess the Discond-VAE models by the two disentanglement metrics (FactorVAE metric \cite{FactorVAE} and MIG \cite{BetaTCVAE}) and the accuracy.

Most of the disentanglement metrics assume that the latent variables are continuous. For example, the $\beta$-VAE metric \cite{BetaVAE} and FactorVAE metric \cite{FactorVAE} measures the degree of disentanglement from the accuracy of the classifier predicting the generative factor based on the variance of each axis of representation. However, our Discond-VAE model and JointVAE adopt the discrete variable to represent the discrete generative factor. Therefore, we evaluate the disentanglement metric on the continuous latent variable based on the continuous generative factors on the CondSprites and dSprites.

Moreover, the CondSprites has class-dependent latent variables. Because each mode of a private variable can represent different variations of the corresponding class, evaluating the disentanglement metric on the entire dataset gives an inappropriate evaluation of the private generative factors. Therefore, we propose a conditional disentanglement evaluation. We define the conditional disentanglement metric as an expectation over discrete variables of the class-wise disentanglement metrics. By the conditional disentanglement metric, we can assess the disentanglement of private factors properly as well as the public factors.
\begin{equation}
    \begin{split}
        \textrm{Conditional Metric} = \mathbb{E}_{p(\cvar)} \left[ \textrm{Metric}(\mathbf{X_{\cvar}})  \right]
    \end{split}
\end{equation}
where $\mathbf{X_{\cvar}}$ denotes the examples from $\mathbf{X}$ with the class $\cvar$.

We evaluate the disentanglement of discrete factors by accuracy. The CondSprites and dSprites have a discrete factor of shape and MNIST has a discrete factor of digit-type. We consider the discrete variable encoder $q(\cvar | \xvar)$ as an unsupervised majority-vote classifier and evaluate its accuracy.

\begin{table*}[t]
\begin{tabular}{l| c cc cc | c cc cc}
\thickhline
\multirow{3}{*}{Method}                  & \multicolumn{5}{c|}{CondSprites}                                                             & \multicolumn{5}{c}{dSprites}                      \\ \cline{3-6} \cline{8-11}
                            & \multirow{2}{*}{Pb, Prv}   & \multicolumn{2}{c}{Cond-FactorVAE(\%)} & \multicolumn{2}{c|}{Cond-MIG}                &  \multirow{2}{*}{Pb, Prv}   & \multicolumn{2}{c}{FactorVAE(\%)} & \multicolumn{2}{c}{MIG}          \\ 
                            &    & Mean (std)               & Best       & Mean (std)     & Best                                              &    & Mean (std)         & Best      & Mean (std)      & Best   \\ \hlineB{2}
\multirow{2}{*}{Joint}      & 10, -  & 74.3 (12.1)             & 87.0     & 0.188 (0.075)   & 0.284                                           & 6, - & \textbf{92.1} (0.2)      & 92.5      & \textbf{0.336} (0.002)   & 0.337                       \\ 
                            & 5, -  & 73.4 (4.1)               & 76.1     & 0.243 (0.041)   & 0.305                                            & 4, -  & \textbf{98.9} (0.4)      & 99.1      & 0.223 (0.022)   & 0.241             \\ \hline
\multirow{4}{*}{\shortstack{Exact \\ (ours)}}     & 10, 3  & \textbf{96.0} (5.9)       & \textbf{100}       & \textbf{0.291} (0.065)   & 0.322      & 6, 2 & 91.6 (0.0)      & 91.6      & 0.309 (0.024)    & 0.338                               \\  
                            & 5, 3  & \textbf{98.5} (2.4)       & \textbf{100}       & \textbf{0.385} (0.124)   & \textbf{0.466}        & 4, 2 & 83.2 (3.7)      & 88.8      & \textbf{0.355} (0.017)   & 0.382             \\ \cline{2-6}
                            & 8, 2  & \textbf{99.8} (0.5)  & \textbf{100}  & \textbf{0.362} (0.028)  &0.388 & 2, 2  & 81.8 (2.2)     & 82.5     & 0.329 (0.065)    & 0.390  \\
                            & 3, 2  & \textbf{95.5} (5.6)  & \textbf{100}  & \textbf{0.267} (0.079) & \textbf{0.363} & --- &       &       &    &  \\\hline
\multirow{4}{*}{\shortstack{Approx \\ (ours)}}     & 10, 3  & 92.7 (8.5)              & \textbf{100}      & 0.201 (0.120)   & \textbf{0.396}                                            & 6, 2 & 90.1 (8.5)     & \textbf{99.8}      & 0.299 (0.065)   & \textbf{0.376}     \\ 
                            & 5, 3  & 97.7 (2.3)               & \textbf{100}     & 0.228 (0.103)   & 0.442                                            & 4, 2  & 92.1 (7.3)      & \textbf{99.8}     & 0.340 (0.039)   & 0.419                               \\  \cline{2-6}
                            & 8, 2  & 92.6 (6.6)               & \textbf{100}      & 0.208 (0.101)   & \textbf{0.402}                                            & 2, 2  & 89.4 (4.9)      & 94.0     & \textbf{0.397} (0.044)   & \textbf{0.454}             \\ 
                            & 3, 2  & 89.0 (10.1)               & 99.8      & 0.206 (0.101)   & 0.339                                           & --- &       &      &    &                                \\ \thickhline
\end{tabular}
\caption{
Disentanglement scores on dSprites and CondSprites. (The higher, the better) \textit{Pb} and \textit{Prv} represent the dimension of the public and private variables. For each JointVAE model, we compare the Discond-VAE models with the same public dimension (upper block) and with the same continuous dimension (lower block). The best scores for each combination of a dataset, dimension, and metric are shown in boldface. On the CondSprites, which has class-dependent and class-independent factors, the Discond-VAE outperforms the JointVAE. Under the unfavorable condition of dSprites, the result shows that the Discond-VAE can adjust the private variable to represent the public factors.
}
\label{table:CondSprites disentanglement metric}
\end{table*}

\subsection{Experiment results on CondSprites} \label{Sec:CondSprites}

\subsubsection{Quantitative Result}

The Discond-VAE-exact model achieves much higher disentanglement scores in both metrics than the JointVAE on the CondSprites in Table \ref{table:CondSprites disentanglement metric}. The Discond-VAE-approx model shows the higher FactorVAE metric and comparable MIG to the JointVAE. The disentanglement results demonstrate that the Discond-VAE can disentangle the private variables. Furthermore, we report the classification accuracy on the CondSprites in Table \ref{table:CondSprites_accuracy}. The Discond-VAE outperforms the JointVAE by a significant margin on the CondSprites.
By disentangling the private variables from the discrete variable, the Discond-VAE can attain a more disentangled discrete representation, which is proved by a higher classification accuracy in Table \ref{table:CondSprites_accuracy}.

\subsection{Experiment results on dSprites}

\begin{table}[!]    
\centering
\begin{tabular}{l c c c c c c}
\thickhline
Method                    & Pb & Prv  & Mean (std)  & Best  \\ \hlineB{2}
\multirow{2}{*}{Joint} & 10 & - & 0.617 (0.068) & 0.720 \\  
                          & 5  & - & 0.599 (0.064) & 0.704 \\ \hline
\multirow{4}{*}{\shortstack{Exact \\ (ours)}}  & 10 & 3 & 0.630 (0.060) & 0.763 \\ 
                          & 5 & 3  & \textbf{0.648} (0.083) & 0.805 \\ \cline{2-5}
                          & 8 & 2  & 0.641 (0.088) & 0.778 \\
                          & 3 & 2  & 0.613 (0.103) &  0.853 \\\hline
\multirow{4}{*}{\shortstack{Approx \\ (ours)}}  & 10 & 3 & \textbf{0.679} (0.121) & \textbf{0.943} \\ 
                          & 5 & 3  & 0.595 (0.088)  & \textbf{0.825}  \\ \cline{2-5}
                          & 8 & 2  &  \textbf{0.677} (0.118) & \textbf{0.946}  \\ 
                          & 3 & 2  & \textbf{0.724} (0.146) & \textbf{0.962}  \\
                          \thickhline
\end{tabular}
\caption{
Unsupervised classification accuracy for CondSprites.
}
\label{table:CondSprites_accuracy}
\end{table}

\begin{table}[!]
\centering
\begin{tabular}{l c c c c}
\thickhline
Method                    & Pb & Prv & Mean (std)   & Best  \\ \hlineB{2}
\multirow{2}{*}{Joint} & 6 & - & $\text{\textbf{0.448}}^* (0.039)$ & $\text{\textbf{0.531}}^*$ \\  
                          & 4 & - & $\text{\textbf{0.440}}^* (0.039)$ & $\text{\textbf{0.541}}^*$ \\ \hline
\multirow{3}{*}{\shortstack{Exact \\ (ours)}}  & 6 & 2 & 0.389 (0.040) & 0.444 \\  
                          & 4 & 2 & 0.369 (0.010)  & 0.381 \\
                          & 2 & 2& 0.351 (0.005)& 0.361\\ \hline
\multirow{3}{*}{\shortstack{Approx \\ (ours)} }  & 6 & 2 &  0.426 (0.044)           & 0.460      \\  
                          & 4 & 2 & 0.434 (0.032)            & 0.449      \\ 
                          & 2 & 2 & \textbf{0.458} (0.012)            & 0.482      \\ 
                          \thickhline

\end{tabular}
\caption{
Unsupervised classification accuracy for dSprites. $^*$ indicates the results from \cite{CascadeVAE}.
}
\label{table:dSprites_Accuracy}
\end{table}

\subsubsection{Quantitative Result}

The dSprites is a synthetic dataset created with five \textit{independent} generative factors. Hence, the probabilistic model of the JointVAE is a more suitable assumption for representing the dSprites compared to that of Discond-VAE. Nevertheless, the Discond-VAE-approx models show a comparable classification accuracy in Table \ref{table:dSprites_Accuracy}, and a similar disentanglement metric of the continuous variables in Table \ref{table:CondSprites disentanglement metric}.

For the case of the two-dimensional public variable, both of the Discond-VAE models show a relatively low FactorVAE score compared to the other models of the same type. In fact, the two-dimensional public variable is insufficient to model four independent continuous factors of dSprites. Nevertheless, the FactorVAE scores are higher than the theoretical limit with two public variables only. This result implies that Discond-VAE can adapt the private variables to represent public generative factors. Considering the CondSprites and dSprites results, the Discond-VAE-exact disentangles the private factors better but has less flexibility to adapt the private variables.


\subsection{Experiment results on MNIST}

\subsubsection{Accuracy and Negative Log-likelihood(NLL)}

\begin{figure}[!]
     \centering
     \begin{subfigure}{0.2\textwidth}
         \centering
         \includegraphics[width=0.8\linewidth]{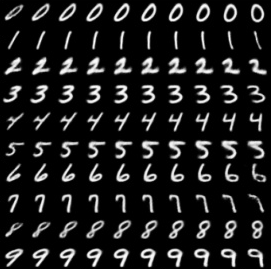}
         \caption{Angle}
         \label{fig:traversal_mnist_angle}
     \end{subfigure}
     \begin{subfigure}{0.2\textwidth}
         \centering
         \includegraphics[width=0.8\linewidth]{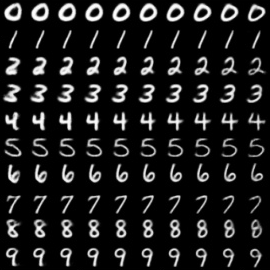}
         \caption{Slant}
         \label{fig:traversal_mnist_slant}
     \end{subfigure}
     \hfill
     
     \begin{subfigure}{0.2\textwidth}
         \centering
         \includegraphics[width=0.8\linewidth]{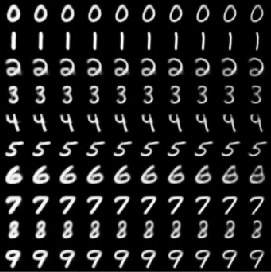}
         \caption{Thickness}
         \label{fig:traversal_mnist_thick}
     \end{subfigure}
    \begin{subfigure}{0.2\textwidth}
         \centering
         \includegraphics[width=0.8\linewidth]{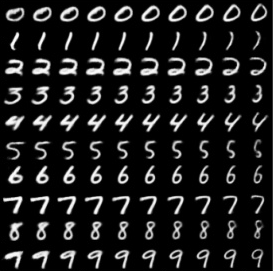}
         \caption{Width}
         \label{fig:traversal_mnist_width}
     \end{subfigure}
     \hfill
    \caption{Public variable traversal on MNIST. Each subfigure corresponds to a different public variable and each row shows the latent traversal of a test example. Discond-VAE encodes public generative factors of MNIST such as Angle, Slant, Thickness, and Width.}
    \label{fig:traversal_mnist_public}
\end{figure}

Table \ref{table:MNIST_Accuracy} shows the unsupervised classification accuracy and NLL scores of each model on the MNIST. The NLL of each model is evaluated by the standard importance weighted sampling strategy \cite{iwae} with 100 samples. Both types of Discond-VAE show a similar or better accuracy and NLL scores compared to the JointVAE while representing the additional private variables. The result indicates that introducing the private variable to learn the intra-class variation provides a capability to disentangle the discrete factors and learn the data distribution better. Both types of Discond-VAE models with two-dimensional public variable show lower NLL scores compared to the other cases. We suspect this is because a two-dimensional public variable is insufficient to model the major public variations of MNIST such as Angle and Thickness in Fig \ref{fig:traversal_mnist_public}.



\begin{table}[!]
\centering
\begin{tabular}{l c c c c c}
\thickhline
Method                    & Pb&Prv  & Mean (std)   & Best   & NLL  \\ \hlineB{2}
\multirow{2}{*}{JointVAE} & 10& - & 0.686 (0.092) & \textbf{0.809} & 152.5 \\ 
                          & 4& -   & 0.708 (0.059) & 0.792 & 153.3\\ \hline
\multirow{4}{*}{\shortstack{Exact \\ (ours)}}  & 10&3  & 0.686 (0.078) &  0.807 & \textbf{145.4}\\  
                          & 4&3  &   \textbf{0.722} (0.099) & \textbf{0.876}  & \textbf{144.7}\\ \cline{2-6}
                          & 8&2  & 0.704(0.064)  &0.828 & \textbf{147.3} \\
                          & 2&2  & 0.712(0.038) & 0.765 & \textbf{152.6} \\ \hline
\multirow{4}{*}{\shortstack{Approx \\ (ours)}}  & 10&3 &\textbf{0.723} (0.069)        & 0.804  & 153.4    \\ 
                          & 4&3  &     0.718 (0.052)        & 0.792 & 153.0    \\ \cline{2-6} 
                          & 8&2  &  \textbf{0.755} (0.076) &  \textbf{0.830} & 150.3 \\
                          & 2&2  & \textbf{0.716} (0.075) & \textbf{0.832}  & 163.0 \\ \thickhline
\end{tabular}
\caption{
Unsupervised classification accuracy and negative log-likelihood(NLL) for MNIST.
}
\label{table:MNIST_Accuracy}
\end{table}

\subsubsection{Latent Traversal}
We observed the latent traversals of the Discond-VAE to evaluate the disentanglement property qualitatively. For the continuous latent variables, each row corresponds to the latent traversals of an axis over a given example. For the discrete variable, each row shows the one-hot traversal of the discrete variable.

\begin{figure}[!]
     \centering
     \begin{subfigure}{0.2\textwidth}
         \centering
         \includegraphics[width=0.98\linewidth]{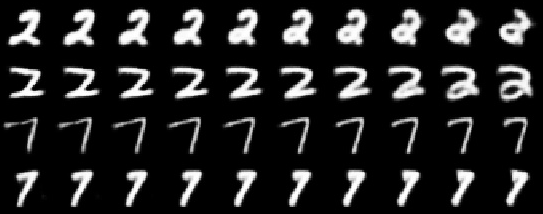}
         \caption{Ring of 2}
         \label{fig:traversal_mnist_2}
     \end{subfigure}
     \begin{subfigure}{0.2\textwidth}
         \centering
         \includegraphics[width=0.98\linewidth]{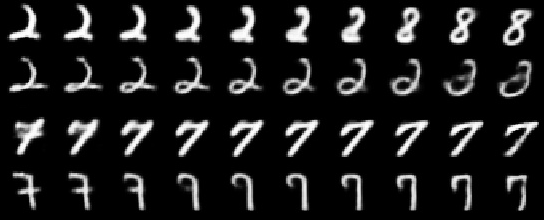}
         \caption{Center-stroke of 7}
         \label{fig:traversal_mnist_7}
     \end{subfigure}
     \hfill
     
     \begin{subfigure}{0.4\textwidth}
         \centering
         \includegraphics[width=0.5\linewidth]{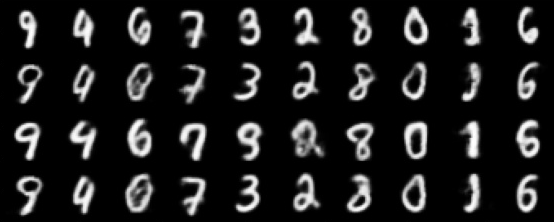}
         \caption{Discrete variable traversal}
         \label{fig:traversal_mnist_disc}
     \end{subfigure}
     \hfill
    \caption{Private and Discrete variable traversal on MNIST. Fig \ref{fig:traversal_mnist_2} and \ref{fig:traversal_mnist_7} shows private variable traversal. Two private variables encode private generative factors such as Ring of 2 and Center-stroke of 7 to the corresponding classes and irrelevant variations to the non-corresponding classes.
    }
    \label{fig:traversal_mnist_private_discrete}
\end{figure}

The Discond-VAE shows a smooth variation in angle, slant, thickness, and width of the decoded images as we traverse the public variable in Fig \ref{fig:traversal_mnist_public}. In the discrete variable traversal (Fig \ref{fig:traversal_mnist_disc}), we can observe a transition in digit-type of given examples. These results demonstrate that the Discond-VAE can disentangle the public and discrete generative factors from the MNIST.

Moreover, the Discond-VAE discovers the class-specific variation of the digit-type 2 and 7. The Fig \ref{fig:traversal_mnist_2} and \ref{fig:traversal_mnist_7} shows the private variable traversal of the Discond-VAE. Each private variable in Fig \ref{fig:traversal_mnist_2} and \ref{fig:traversal_mnist_7} represents the ring of digit-type 2 and the center-stroke of digit-type 7, respectively. Since these two variations are exclusive to each class, the latent traversals of these two variables show relatively minor or irrelevant variations to the other class images.

\subsection{Experiment results on CelebA}
\subsubsection{Latent Traversal}



\begin{figure}[!]
     \centering
     \begin{subfigure}{0.4\textwidth}
         \centering
         \includegraphics[width=1\linewidth]{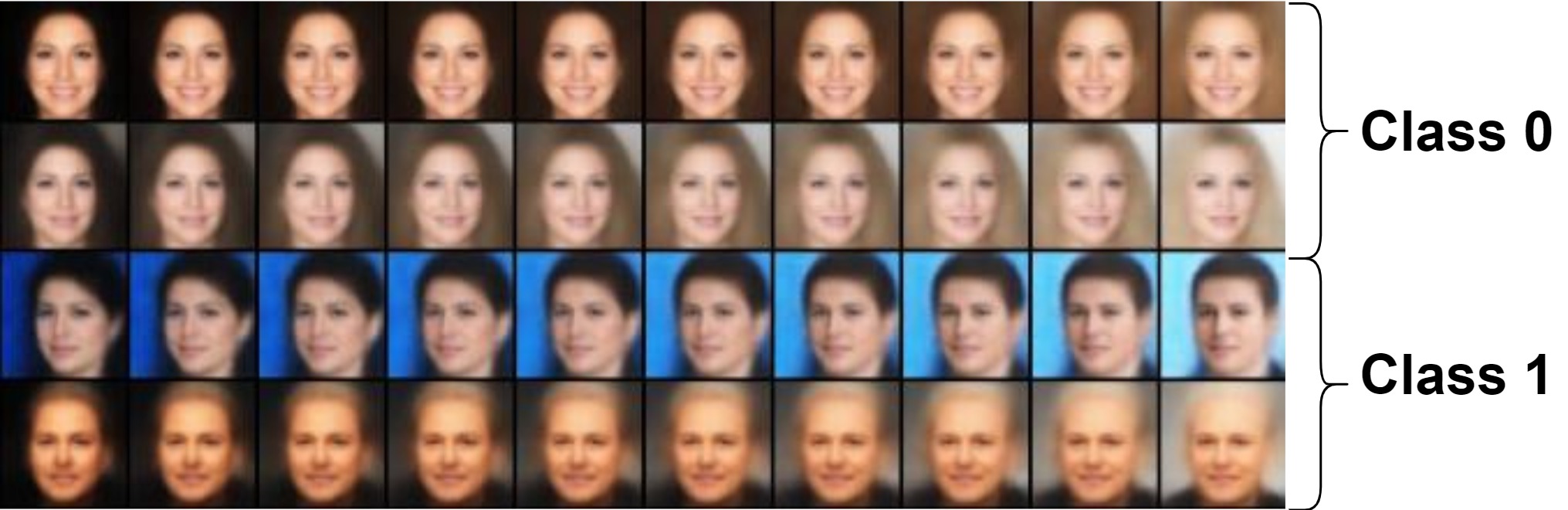}
         \caption{Public - Brightness}
         \label{fig:celeba_public_brightness_crop}
     \end{subfigure}
     \begin{subfigure}{0.4\textwidth}
         \centering
         \includegraphics[width=1\linewidth]{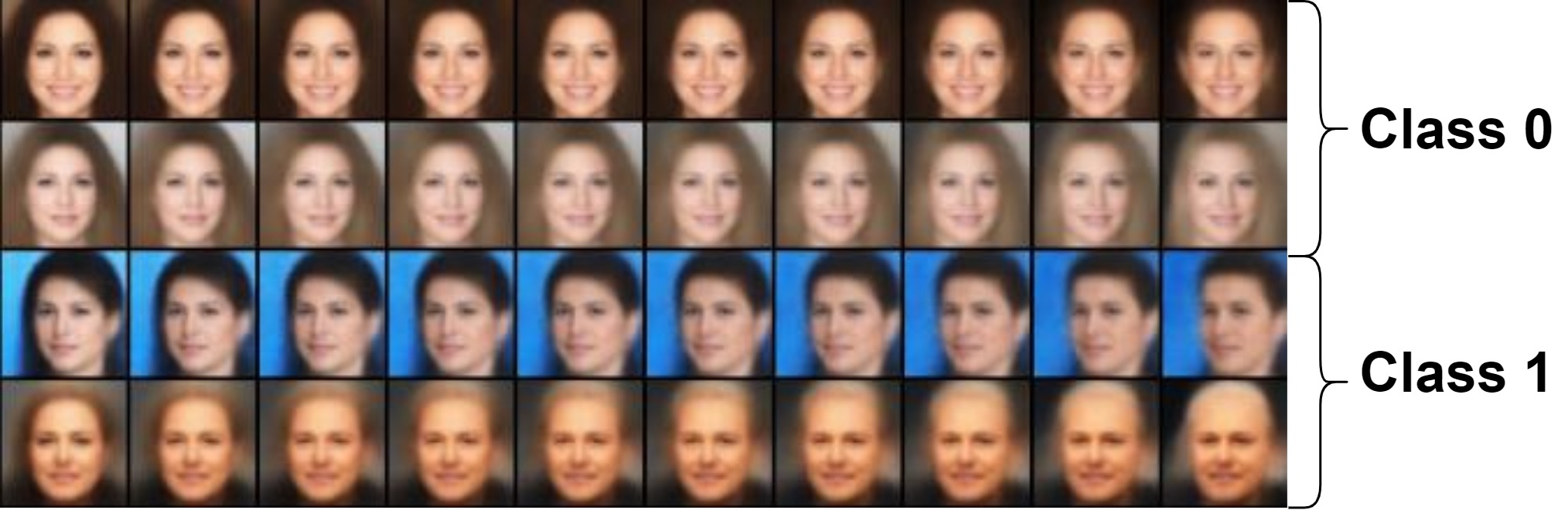}
         \caption{Private - Gender (Class 1)}
         \label{fig:celeba_prv_gender_crop}
     \end{subfigure}
     \hfill
     
    \caption{Public and Private variable traversal on CelebA. The public variable represents the Brightness for the both classes. The private variable represents the Gender for the class 1 (Female $\rightarrow$ Male) but irrelevant variations for the class 0.
    }
    \label{fig:celeba_traversal}
\end{figure}

To further test the generalizability of Discond-VAE, we observed the latent traversals on the CelebA. Fig \ref{fig:celeba_traversal} shows that Discond-VAE can disentangle the public and private generative factors on the more challenging domain of CelebA. The top two rows and the bottom two rows in Fig \ref{fig:celeba_traversal} represent latent traversals of two examples from the two different classes. The traversal of the public variable (Fig \ref{fig:celeba_public_brightness_crop}) generates variation in brightness for the both classes. For the private variable(Fig \ref{fig:celeba_prv_gender_crop}), the Gender traversal is observed for the class 1, but such change is not observed in the class 0. (See appendix for the more public traversals representing the background and face-width.)

\section{Related Works}
Extracting disentangled features from data without supervision is an important task for representation learning. \cite{BengioRepresentation} Several VAE variants adopting continuous latent variables are proposed to obtain more disentangled representations. For example, $\beta$-VAE \cite{BetaVAE} increases a disentangling pressure of VAE by increasing the weight of the KL divergence between the variational posterior $q(\zvar | \xvar)$ and the prior $p(\zvar)$. 
The KL divergence regularizer of $\beta$-VAE penalizes not only the total correlation (TC) of the aggregate posterior $q(\zvar)$, which induces the factorized posterior, but also the mutual information between the data and the latent variables. Since penalizing the mutual information is detrimental to extract meaningful features, several works proposed penalizing TC only in various ways. (\eg auxiliary discriminator in FactorVAE \cite{FactorVAE}, mini-batch weighted sampling in $\beta$-TCVAE \cite{BetaTCVAE}, and covariance matching in DIP-VAE \cite{DIPVAE})
In addition, Bayes-Factor-VAE \cite{bayesfactor} suggests dividing the continuous variables into the relevant variable and nuisance variable. Bayes-Factor-VAE promotes disentangled features by introducing hyper-priors on the variances of Gaussian prior. HFVAE \cite{structured} uses a two-level hierarchical objective to control the independence between groups of variables and the independence between each variable in the same group.

Recently, a number of works to model the intrinsic discreteness of the real-world data are proposed. Some of these works proposed representing the discrete variable by modeling the continuous variable as a multimodal distribution or tree-structured model. GMVAE \cite{bayesruleVAE} represents continuous variables as a Gaussian Mixture and infers the discrete variable by Bayes rule. CascadeVAE \cite{CascadeVAE} proposed an iterative optimization method to minimize the TC of continuous variables and an alternating optimization method between discrete and continuous variables to train the model. CascadeVAE infers the discrete variable via the inner maximization step over the discrete variables \cite{CascadeVAE_opt}. Moreover, \cite{nCRPVAE} and LTVAE \cite{LTVAE} encode the latent variable as a tree-structured model and learns the tree structure from the data themselves. \cite{nCRPVAE} employs nested Chinese Restaurant Process \cite{nCRP} to accommodate a hierarchical prior to the data. LTVAE adjusts a tree-structure of the latent variable via the EM algorithm. Furthermore, VQ-VAE \cite{VQVAE} and VQ-VAE-2 \cite{VQVAE2} proposed a discrete code representation of continuous variable by introducing a nearest neighbor loop-up.

By contrast, JointVAE \cite{JointVAE} and InfoCatVAE \cite{infocatvae} have an explicit encoder to encode discrete variable. JointVAE \cite{JointVAE} proposed a method of jointly training continuous and discrete variables. However, JointVAE has a limitation of assuming that the discrete and continuous variables are independent of each other. By introducing the private latent variables, our Discond-VAE can represent the dependent discrete and continuous variable structure.
InfoCatVAE \cite{infocatvae} proposed encoding a conditional distribution of continuous variable $q(\wvar | \xvar, \cvar)$. In this respect, the private variable of our proposed Discond-VAE and InfoCatVAE have a similar probabilistic formulation. However, InfoCatVAE adopts the axis-division of Gaussian to separate the meaningful variables of each class. The axis-division strategy requires each subdivision to encode a certain latent variable even for the irrelevant classes while the mode-division strategy of our Discond-VAE does not. Therefore, Discond-VAE can promote more disentangled representation compared to InfoCatVAE, even only with the private variable.

\section{Conclusion}
We proposed Discond-VAE for learning the public and private continuous generative factors and the discrete generative factor from the data. We developed a probabilistic framework and the learning objective for the Discond-VAE, and suggested two implementations of the framework according to how the private variable is addressed. Also, we proposed the CondSprites dataset to evaluate the disentanglement capacity for the class-dependent generative factors of a model. Then, we evaluated both types of the Discond-VAE model on the CondSprites, dSprites, MNIST, and CelebA. The experiment results prove that the Discond-VAE model can disentangle the class-dependent and class-independent factors in an unsupervised manner. Moreover, the Discond-VAE shows a moderate degree of disentanglement even on the dSprites which has only independent generative factors.

{\small
\bibliographystyle{ieee_fullname}
\bibliography{egbib}
}

\clearpage
\appendix
\section{KL divergence of Private and Discrete Variables}
The KL divergence regularizer of the private and discrete variables can be decomposed as the following Eq \ref{sup:kl_prv}. Eq \ref{sup:kl_prv_1} can be interpreted as the expectation of the mode-wise regularizer of the private variable. Eq \ref{sup:kl_prv_2} represents the regularizer of the discrete variable.

\begin{align} \label{sup:kl_prv}
    D_{KL}(q_{\phi}(\mathbf{w}, & \mathbf{c} \mid \xvar) \mid\mid p(\mathbf{w}, \mathbf{c})) \\
    = \mathbb{E}_{q_{\phi}(\mathbf{c}\mid \mathbf{x})} &\mathbb{E}_{q_{\phi}(\mathbf{w}\mid \mathbf{x}, \mathbf{c})} [ \log \frac{q_{\phi}(\mathbf{w}\mid \mathbf{x}, \mathbf{c}) \cdot q_{\phi}(\mathbf{c}\mid \mathbf{x})}{p(\mathbf{w} \mid \mathbf{c}) \cdot p(\mathbf{c})  } ] \\
    = \mathbb{E}_{q_{\phi}(\mathbf{c} \mid \mathbf{x})} &\left [ \mathbb{E}_{q_{\phi}(\mathbf{w}\mid \mathbf{x}, \mathbf{c})} [ \log \frac{q_{\phi}(\mathbf{w}\mid \mathbf{x}, \mathbf{c}) }{p(\mathbf{w} \mid \mathbf{c})} ] \right ]\\ 
    & \qquad + \mathbb{E}_{q_{\phi}(\mathbf{c} \mid \mathbf{x})} \left [ \log \frac{q_{\phi}(\mathbf{c}\mid \mathbf{x})}{p(\mathbf{c})  }  \right] \\
    = \mathbb{E}_{q_{\phi}(\mathbf{c} \mid \mathbf{x})} &\left [ \mathbb{E}_{q_{\phi}(\mathbf{w}\mid \mathbf{x}, \mathbf{c})} [ \log \frac{q_{\phi}(\mathbf{w}\mid \mathbf{x}, \mathbf{c}) }{p(\mathbf{w} \mid \mathbf{c})} ] \right] \\
    & \qquad + D_{KL}(q_{\phi}(\mathbf{c} \mid \xvar) \mid\mid p(\mathbf{c})) \\ 
    = \mathbb{E}_{q_{\phi}(\mathbf{c} \mid \mathbf{x})} &\left [ D_{KL} (q_{\phi}(\wvar \mid \xvar, \cvar \mid \mid p(\wvar \mid \cvar) )  \right] \label{sup:kl_prv_1} \\
    & \qquad + D_{KL}(q_{\phi}(\mathbf{c} \mid \xvar) \mid\mid p(\mathbf{c})) \label{sup:kl_prv_2}
\end{align}

\section{Mixture prior Optimization}
The Discond-VAE models the private variable as the following.
\begin{align} 
    p(\wvar \mid \cvar) &= \prod_{i} p(\wvar \mid \cvar = e_{i})^{c_i}  \\
    &= \prod_{i} \mathcal{N}\left(\mu_i, I\right)^{c_i} \\
    q_{\phi}(\wvar \mid \xvar, \cvar) &= \prod_{i} q_{\phi}(\wvar \mid \xvar, \cvar=e_{i})^{c_i} \\
                        &= \prod_{i} \mathcal{N}\left(\mu_{i}(\xvar, \cvar), \Sigma_{i}(\xvar,\cvar)\right)^{c_i}
\end{align}
The Discond-VAE sets the mean vector $\mu_{i}$ of Gaussian Mixture prior by the zero for Discond-VAE-exact and by the random samples for Discond-VAE-approx. These initialization methods do not reflect the semantic relationship of discrete generative factors.  Thus, we tried the EM algorithm or Warm-up approaches to optimize the $\mu_{i}$.

The KL divergence term of private variable $\wvar$ is expressed as
\begin{equation}
    \begin{split}
        &\mathbb{E}_{q_{\phi}(\mathbf{c}\mid \mathbf{x})} 
        \left [ D_{KL} (q_{\phi}(\wvar \mid \xvar , \cvar) \mid\mid p(\wvar \mid \cvar)) \right] \\
        = &\sum_{i=1}^{d} \alpha_{i} \cdot D_{KL} (q_{\phi}(\wvar \mid \xvar , \cvar = e_{i}) \mid\mid p(\wvar \mid \cvar = e_{i})) 
    \end{split}
\end{equation}
Since $ q_{\phi}(\wvar \mid \xvar, \cvar=e_{i})$ is a factorized Gaussian distribution, the KL divergence of each mode becomes
\begin{equation}
    \begin{split}
        & D_{KL} (q_{\phi}(\wvar \mid \xvar , \cvar) \mid\mid p(\wvar \mid \cvar))\\
        = &\frac{1}{2} \cdot \sum_{j} \left( \left(\mu_{i,j}(\xvar,\cvar) - \mu_{i,j}\right)^{2} + \sigma_{i,j}^{2} - \log(\sigma_{i,j}^2) - 1
        \right)
    \end{split}
\end{equation}
where $j$ indexes the dimension of the private variable. Therefore, we can obtain a closed-form solution $\boldsymbol{\mu^{*}}$ of the optimization problem below over the $\boldsymbol{\mu}$.
\begin{equation}\label{eq:MixturePriorOpt}
    \begin{split}
        \arg\min_{\boldsymbol{\mu}} \,\, &\mathbb{E}_{p(\xvar)} \mathbb{E}_{q_{\phi}(\mathbf{c}\mid \mathbf{x})} 
        \left [ D_{KL} (q_{\phi}(\wvar \mid \xvar , \cvar) \mid\mid p(\wvar \mid \cvar)) \right] \\
        \mu_{i,j}^{*} &= \frac{\sum_{n, i} \alpha_{n,i}(x_{n}) \cdot \mu_{i,j}(x_{n})}
        {\sum_{i, j} \alpha_{n,i}(x_{n})}
    \end{split}
\end{equation}
where $n$ indexes the sample $x_n$ from training data.

To exploit the semantic features extracted from the model, we updated the mean vector $\boldsymbol{\mu}$ (Eq \ref{eq:MixturePriorOpt}) by the EM-like or Warm-up approaches. The Warm-up approach updates the mean vector \textit{once} right after the $10\%$ of entire training epochs. The EM approach updates the mean vector \textit{for every} $10\%$ of the entire training epochs. However, these approaches showed the inferior performance to the initialize-and-fix methods and often did not converge.

\begin{table*}[!] 
\centering
\begin{tabular}{c c c c c}
\multicolumn{3}{c}{\textbf{Encoder} } & \multicolumn{2}{c}{\textbf{Decoder} } \\ \thickhline
\multicolumn{3}{c}{1$\times$32 Conv 4$\times$4, stride 2} &\textbf{Public} & \textbf{Private}\\
\multicolumn{3}{c}{\begin{tabular}[c]{@{}l@{}} (32$\times$32 Conv 4$\times4$, stride 2)\end{tabular}} &  Linear Pb $\times$ 128 & Linear ($d$ * Pr + $d$) $\times$ 128 \\
\multicolumn{3}{c}{32$\times$64 Conv 4$\times$4, stride 2} & \multicolumn{2}{c}{\begin{tabular}[c]{@{}l@{}}Linear 256 $\times$ (64 * 4 * 4)\end{tabular}}\\
\multicolumn{3}{c}{64$\times$64 Conv 4$\times$4, stride 2} &  \multicolumn{2}{c}{(64 $\times$ 64 Conv Transpose 4$\times$4, stride 2)} \\
\multicolumn{3}{c}{Linear (64 * 4 * 4) $\times$ 256} & \multicolumn{2}{c}{64 $\times$ 32 Conv Transpose 4$\times$4, stride 2}\\
\textbf{Public} &  \textbf{Private} &   \textbf{Discrete}& \multicolumn{2}{c}{32 $\times$ 32 Conv Transpose 4$\times$4, stride 2}  \\ 
Linear 256 $\times$ Pb &  Linear (256+$d$) $\times$ $d$ * Pr &   Linear 256 $\times$ $d$& \multicolumn{2}{c}{32 $\times$ 1 Conv Transpose 4$\times$4, stride 2}  \\\thickhline
\end{tabular}
\caption{Discond-VAE-exact architecture. For each layer, $a \times b$ at the front represents that the layer has $a$ in-channels and $b$ out-channels. The layers in () are added for the dSprites and CondSprites. $d$ denotes the dimension of the discrete variable. Pr and Pb represent the dimension of the private and public variables, respectively.}
\label{table:Discond_exact_arch}
\end{table*}

\begin{table*}[!] 
\centering
\begin{tabular}{c c c c c}
\multicolumn{3}{c}{\textbf{Encoder} } & \multicolumn{2}{c}{\textbf{Decoder} } \\ \thickhline
\multicolumn{3}{c}{1$\times$32 Conv 4$\times$4, stride 2} & \multicolumn{2}{c}{Linear (Pb + Pr + $d$) $\times 256$} \\
\multicolumn{3}{c}{\begin{tabular}[c]{@{}l@{}} (32$\times$32 Conv 4$\times4$, stride 2)\end{tabular}} & \multicolumn{2}{c}{\begin{tabular}[c]{@{}l@{}}Linear 256 $\times$ (64 * 4 * 4)\end{tabular}} \\
\multicolumn{3}{c}{32$\times$64 Conv 4$\times$4, stride 2} & \multicolumn{2}{c}{(64 $\times$ 64 Conv Transpose 4$\times$4, stride 2)} \\
\multicolumn{3}{c}{64$\times$64 Conv 4$\times$4, stride 2} & \multicolumn{2}{c}{64 $\times$ 32 Conv Transpose 4$\times$4, stride 2} \\
\multicolumn{3}{c}{Linear (64 * 4 * 4) $\times$ 256} & \multicolumn{2}{c}{32 $\times$ 32 Conv Transpose 4$\times$4, stride 2} \\
\begin{tabular}[c]{@{}l@{}} \textbf{Public}\\ Linear 256 $\times$ Pb\end{tabular} & \begin{tabular}[c]{@{}l@{}} \textbf{Private}\\ Linear (256+$d$) $\times$ $d$ * Pr\end{tabular} & \begin{tabular}[c]{@{}l@{}}\textbf{Discrete}\\ Linear 256 $\times$ $d$\end{tabular} & \multicolumn{2}{c}{32 $\times$ 1 Conv Transpose 4$\times$4, stride 2} \\ \thickhline                                      
\end{tabular}
\caption{Discond-VAE-approx architecture. The same notation as in Table \ref{table:Discond_exact_arch} is applied.
}
\label{table:Discond_approx_arch}
\end{table*}

\section{Implementation details}
\subsection{Network Architecture}   
We use similar network architectures as in JointVAE \cite{JointVAE}. The only modification is that the Discond-VAE-exact model embeds the public variable and the private and discrete variables separately. (Decoder of the Table \ref{table:Discond_exact_arch})

\subsection{Training details}
As a reminder, the learning objective of Discond-VAE is expressed as the following.
\begin{equation}
    \begin{split}
        & \max_{\theta, \phi} \mathcal{L}_{\textrm{Cond}}(\theta, \phi) =\mathbb{E}_{q_{\phi}(\mathbf{z}, \mathbf{w}, \mathbf{c} \mid \mathbf{x})}[\log p_\theta (\mathbf{x} \mid\mathbf{z}, \mathbf{w}, \mathbf{c} )] \\
        &- \beta_{\zvar} \left| D_{KL}(q_{\phi}(\mathbf{z} \mid \mathbf{x}) \mid\mid p(\mathbf{z}) )  - C_{\zvar} \right|\\
        &- \beta_{\wvar} \left| \mathbb{E}_{q_{\phi}(\mathbf{c}\mid \mathbf{x})} 
        \left [ D_{KL} (q_{\phi}(\wvar \mid \xvar , \cvar) \mid\mid p(\wvar \mid \cvar)) \right]  - C_{\wvar} \right|\\
        & -\beta_{\cvar} \left| D_{KL}(q_{\phi}(\mathbf{c} \mid \xvar) \mid\mid p(\mathbf{c})) - C_{\cvar} \right|
    \end{split}
\end{equation}
For both Discond-VAE models, we apply the linear scheduling of capacity \cite{VAECapacity} as in the JointVAE. Each capacity $C_{\zvar}, C_{\wvar}, C_{\cvar}$ is linearly increased from 0 to $C$ in the iteration hyperparameters. For the JointVAE, we applied the same hyperparameters as in \cite{JointVAE} for the MNIST and dSprites. All models in the paper are optimized by Adam optimizer. The parameters for the Adam optimizer are betas=(0.9, 0.999), eps=$1e^{-8}$ with no weight decay, which is the default setting in the PyTorch library.

\begin{table*}[htbp!]
\centering
\begin{tabular}{c|cccc|ccc|cccc}
 &  \multicolumn{4}{c|}{CondSprites} & \multicolumn{3}{c|}{dSprites} & \multicolumn{4}{c}{MNIST} \\ \thickhline
 $Pb$& 10     & 8      & 5      & 3     & 6        & 4        & 2       & 10    & 8     & 4     & 2     \\
 $Pr$& 3      & 2      & 3      & 2     & 2        & 2        & 2       & 3     & 2     & 3     & 2     \\
 $d$& 2      & 2      & 2      & 2     & 3        & 3        & 3       & 10    & 10    & 10    & 10    \\
 $\beta _ {\zvar}$& 30     & 30     & 30     & 30    & 200      & 100      & 200     & 25    & 25    & 25    & 25    \\
 $\beta _ {\wvar}$& 30     & 30     & 30     & 30    & 200      & 100      & 200     & 25    & 25    & 25    & 25    \\
 $\beta _ {\cvar}$& 30     & 40     & 30     & 40    & 200      & 100      & 200     & 5     & 2.5   & 5     & 3     \\
 $C_ {\zvar}$& 30     & 30     & 30     & 30    & 20       & 20       & 20      & 5     & 5     & 5     & 5     \\
 $C_ {\wvar}$& 30     & 30     & 30     & 30    & 20       & 20       & 20      & 5     & 5     & 5     & 5     \\
  $C_ {\cvar}$& 5      & 5      & 5      & 10    & 1.1      & 1.1      & 1.1     & 25    & 25    & 25    & 25    \\
 iteration & 25000  & 25000  & 25000  & 25000 & 300000   & 300000   & 300000  & 25000 & 25000 & 25000 & 25000   \\ \thickhline
\end{tabular}
\caption{Hyperparameters of DiscondVAE-Exact. Capacity increased linearly from 0 to $C$ in an iteration.  }
\label{table:Discond_exact_hyper}
\end{table*}

\begin{table*}[htbp!]
\centering
\begin{tabular}{c|cccc|ccc|cccc}
 & \multicolumn{4}{c|}{CondSprites}            & \multicolumn{3}{c|}{dSprites} & \multicolumn{4}{c}{MNIST}     \\ \thickhline
 $Pb$& 10 & 8 & 5 & 3                   & 6 & 4 & 2                         & 10 & 8 & 4 & 2 \\
 $Pr$& 3 & 2 &  3 & 2                    &  2 & 2 & 2                       & 3 & 2 & 3 & 2 \\
 $d$& 2 & 2 & 2 & 2                 & 3 & 3 & 3                             & 10 & 10 & 10 & 10\\
 $\beta _ {\zvar}$&10&10&10&20   & 20 &20 & 20                              & 30 & 10 & 20 & 10\\
 $\beta _ {\wvar}$&20&20&20&40   & 40 &40 & 40                              & 60 & 20 & 40 & 20\\
 $\beta _ {\cvar}$&20&20&20&40   & 40 &40 & 40                              & 60 & 20 & 40 & 20\\
 $C_ {\zvar}$&20&20 &20&10        &10 &10 & 10                              & 10 & 10 & 10 & 10\\
 $C_ {\wvar}$&20 &20 &20 &10        &10 &10 & 10                            & 10 & 10 & 10 & 10\\
 $C_ {\cvar}$&5 &5 &5 &5        &5 & 5  & 5                                 & 10 & 10 & 5 & 10\\
 iteration & 25000 & 25000 & 25000 & 25000 & 300000 & 300000 &300000        &25000 &25000 &25000 &25000 \\ \thickhline
\end{tabular}
\caption{Hyperparameters of DiscondVAE-approx. Capacity increased linearly from 0 to $C$ in an iteration.  }
\label{table:Discond_approx_hyper}
\end{table*}

\begin{table*}[htbp!]
\centering
\begin{tabular}{c|cc|cc|cc}
 & \multicolumn{2}{c|}{CondSprites} & \multicolumn{2}{c|}{dSprites} & \multicolumn{2}{c}{MNIST} \\ \thickhline
 $Pb$& 10           & 5           & 6             & 4             & 10              & 4              \\
 $d$& 2            & 2           & 3             & 3             & 10              & 10             \\
 $\beta _ {\zvar}$& 30           & 30          & 150           & 150           & 30              & 30             \\
 $\beta _ {\cvar}$& 30           & 30          & 150           & 150           & 30              & 30             \\
 $C_ {\zvar}$& 30           & 30          & 40            & 40            & 5               & 5  \\
  $C_ {\cvar}$& 5 & 5 &1.1&1.1&5&5\\
 iteration & 25000 & 25000 & 300000 & 300000 & 25000 & 25000 \\ \thickhline
\end{tabular}
\caption{Hyperparameters of JointVAE. Capacity increased linearly from 0 to $C$ in an iteration.  }
\label{table:Joint_hyper}
\end{table*}

\subsection{Discond-VAE-exact Hyperparameters}
\subsubsection{MNIST}
\begin{itemize}
  \item Epochs: 100
  \item Batch size: 64 
  \item Optimizer: Adam with learning rate 5e-4
\end{itemize}

\subsubsection{dSprites}
\begin{itemize}
  \item Epochs: 30
  \item Batch size: 64 
  \item  Optimizer: Adam with learning rate 5e-4
\end{itemize}

\subsubsection{CondSprites}
\begin{itemize}
  \item Epochs: 200
  \item Batch size: 64 
  \item  Optimizer: Adam with learning rate 5e-4
\end{itemize}
All other hyperparameters in Table \ref{table:Discond_exact_hyper}.

\subsubsection{Hyperparameters Search range}
\begin{itemize}
    \item Learning rate - \{ 5e-4\}
    \item $(\beta_{\zvar} = \beta_{\wvar})$ -  \{5, 10,15,20,25 30, 50,100,200\}
    \item $\beta_{\cvar}$ - \{1,3,5,10,20,30,50,100,200 \}
    \item $(C_{\zvar} = C_{\wvar})$ - \{ 5, 10, 20,30,50 \}
    \item $C_{\cvar}$ - \{ 1,1.1 5, 10,25,50 \}
\end{itemize}

\subsection{Discond-VAE-approx Hyperparameters}
\subsubsection{MNIST}
\begin{itemize}
  \item Epochs: 100
  \item Batch size: 64 
  \item Optimizer: Adam with learning rate 2e-3
\end{itemize}

\subsubsection{dSprites}
\begin{itemize}
  \item Epochs: 20
  \item Batch size: 64 
  \item Optimizer: Adam with learning rate 1e-3
\end{itemize}

\subsubsection{CondSprites}
\begin{itemize}
  \item Epochs: 300
  \item Batch size: 64 
  \item Optimizer: Adam with learning rate 1e-3
\end{itemize}
All other hyperparameters in Table \ref{table:Discond_approx_hyper}.

\subsubsection{Hyperparameters Search range}
\begin{itemize}
    \item Learning rate - \{ 1e-3, 2e-3 \}
    \item $\beta_{\zvar}$ -  \{10, 20, 30, 50\}
    \item $\beta_{\zvar} : (\beta_{\wvar} = \beta_{\cvar})$ - \{ 2:1, 1:1, 1:2 \}
    \item $(C_{\zvar} = C_{\wvar})$ - \{ 5, 10, 20 \}
    \item $C_{\cvar}$ - \{ 1, 5, 10 \}
\end{itemize}

\subsection{JointVAE Hyperparameters}
\subsubsection{MNIST}
\begin{itemize}
  \item Epochs: 100
  \item Batch size: 64 
  \item Optimizer: Adam with learning rate 5e-4
\end{itemize}

\subsubsection{dSprites}
\begin{itemize}
  \item Epochs: 30
  \item Batch size: 64 
  \item  Optimizer: Adam with learning rate 5e-4
\end{itemize}

\subsubsection{CondSprites}
\begin{itemize}
  \item Epochs: 200
  \item Batch size: 64 
  \item  Optimizer: Adam with learning rate 5e-4
\end{itemize}
All other hyperparameters in Table \ref{table:Joint_hyper}.

\section{CondSprites Details}
The CondSprites is a subset of the dSprites \cite{dSprites} designed to model the dependence between the continuous and discrete generative factors. We removed the Heart shape to maintain a reasonable amount of examples. Since the Square images do not vary in the $y$-position in Table \ref{table:CondSprites_dataset}, we fix the Square images at the center of $y$-axis. In other words, the generative factor for the $y$-position of Square images in the CondSprites is fixed to 16 (Center of the range(0, 32)). Likewise, the generative factor for the $x$-position of every Ellipse image is fixed to 16. The total number of CondSprites examples is $15360 = 6(\text{Scale}) * 40 (\text{Orientation}) * (32(x\text{ for Squares}) + 32(y\text{ for Ellipses}))$.

\begin{table}[!]
\centering
\begin{tabular}{l c c}
\thickhline
\multirow{2}{*}{Generative factors}      & \multicolumn{2}{c}{Shape} \\ \cline{2-3}
                        & Square & Ellipse\\ \hlineB{2}
Scale                   & \checkmark     & \checkmark     \\ 
Orientation             & \checkmark     & \checkmark     \\  
Position X              & \checkmark     &      \\  
Position Y              &      & \checkmark     \\ \thickhline
\end{tabular}
\caption{
Generative factors of the CondSprites.
}
\end{table}

\begin{figure}[!]
     \centering
     \begin{subfigure}{0.2\textwidth}
         \centering
         \includegraphics[width=1\linewidth]{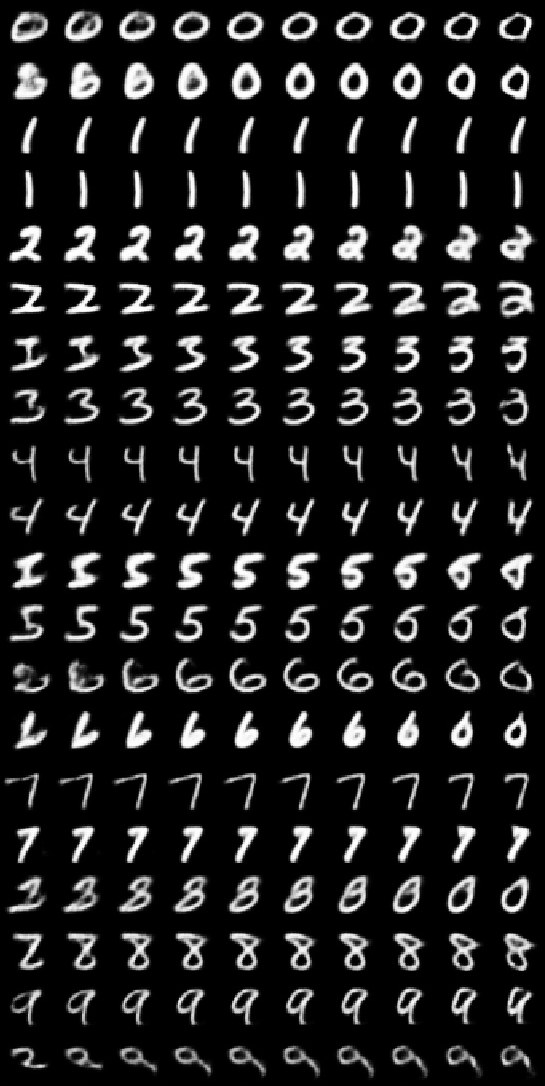}
         \caption{Ring of 2}
         \label{fig:traversal_mnist_prv_full_2}
     \end{subfigure}
     \,\,
     \begin{subfigure}{0.2\textwidth}
         \centering
         \includegraphics[width=1\linewidth]{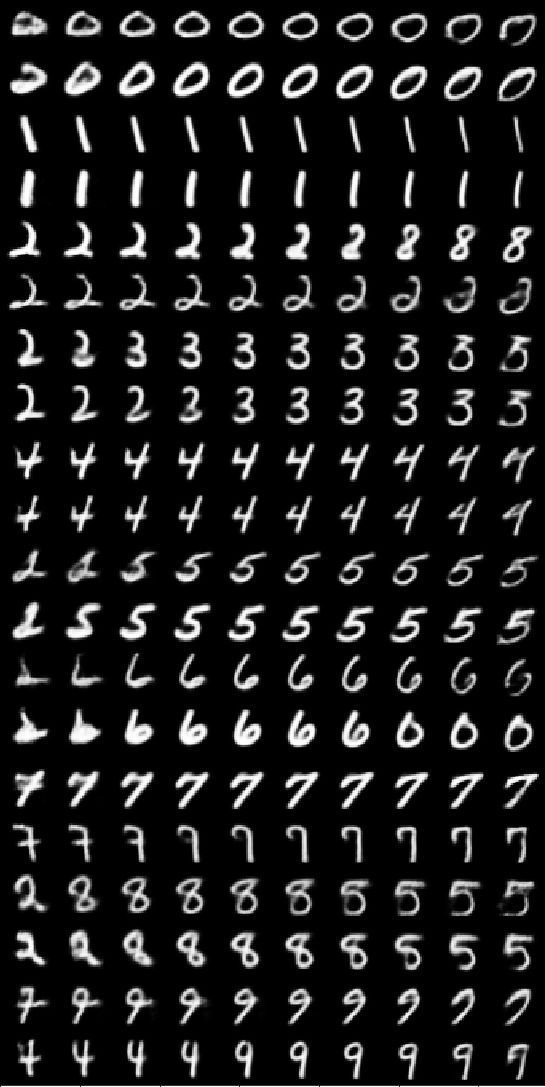}
         \caption{Center-stroke of 7}
         \label{fig:traversal_mnist_prv_full_7}
     \end{subfigure}
     \caption{Private variable traversals of all classes on MNIST. Two private variables represent private generative factors of the Ring of 2 and Center-stroke of 7. Each private variable represents relatively minor or irrelevant variations to the non-corresponding classes.}
     \label{fig:traversal_mnist_prv_full}
\end{figure}

\begin{figure}[!]
     \centering
     \begin{subfigure}{0.4\textwidth}
         \centering
         \includegraphics[width=1\linewidth]{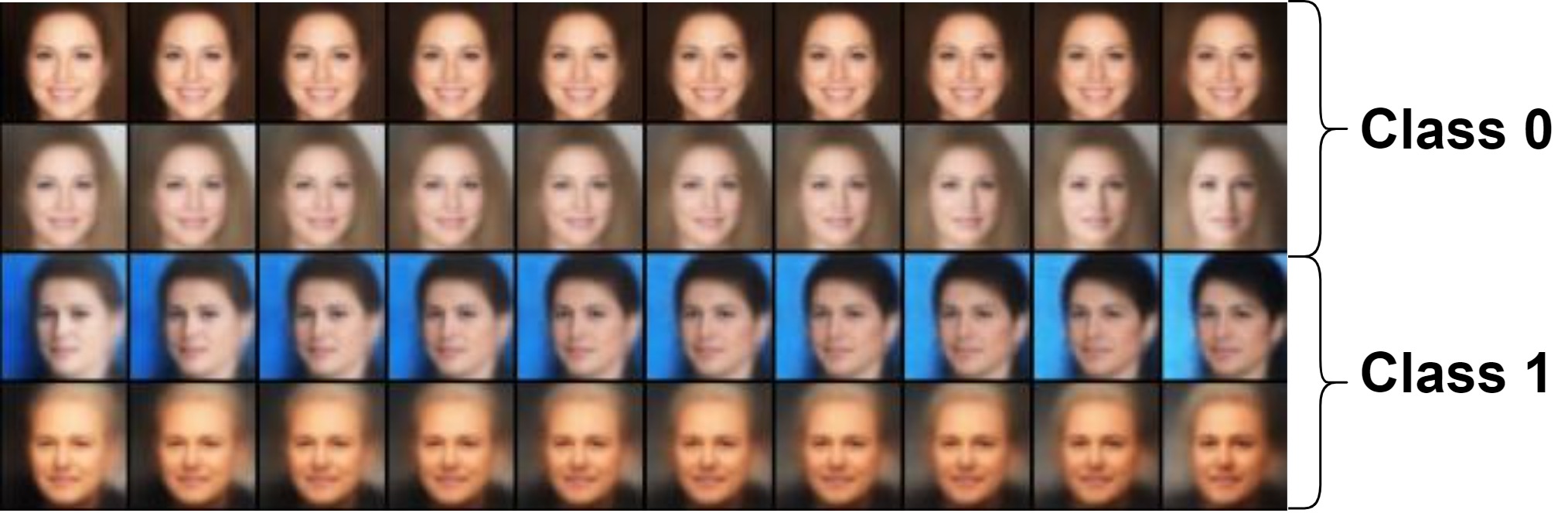}
         \caption{Background}
         \label{fig:celeba_pub_background}
     \end{subfigure}
     \,\,
     \begin{subfigure}{0.4\textwidth}
         \centering
         \includegraphics[width=1\linewidth]{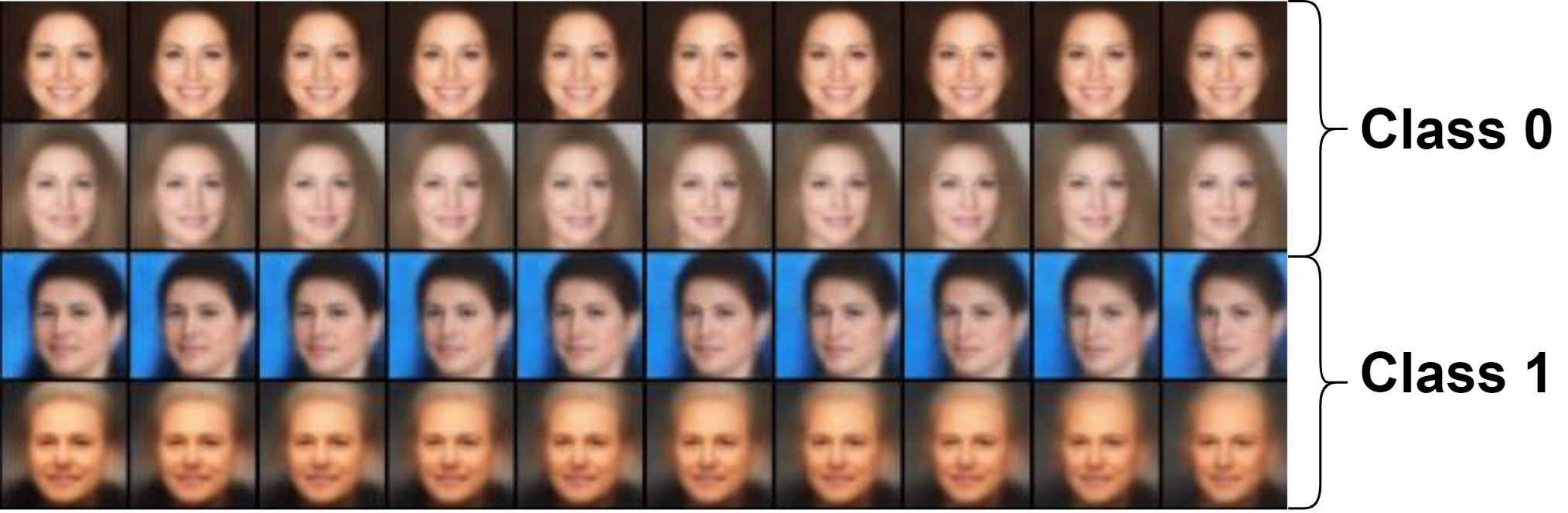}
         \caption{Face-width}
         \label{fig:celeba_pub_face_width}
     \end{subfigure}
     \caption{Public variable traversals on CelebA. The public variables represent the Background and Face-width regardless of the classes.}
     \label{fig:celeba_pub_additional}
\end{figure}

\section{Traversal}
Fig \ref{fig:traversal_mnist_prv_full} shows the private variable traversals of all classes in MNIST \cite{MNIST}. Since the Ring of 2 and Center-stroke of 7 are exclusive intra-class variations of digit-type 2 and 7, latent traversals on the other classes show relatively minor or irrelevant variations. Note that the private variable traversal of an image shows the latent traversal of the corresponding mode in the Mixture of Gaussian.

Fig \ref{fig:celeba_pub_additional} shows the additional public variable traversals in CelebA \cite{CelebA}. The top two rows and the bottom two rows show latent traversals of two examples from the two classes. Since the public variable is independent of the class, each public variable encodes the same variation for the classes. In Fig \ref{fig:celeba_pub_background}, as we traverse the public variable, the background changes from left dark, right bright to left bright, right dark. In Fig \ref{fig:celeba_pub_face_width}, as we traverse the public variable, the face changes from wide to narrow.

\end{document}